\documentclass[letterpaper]{article}
\usepackage[preprint]{aaai2027}
\usepackage[hyphens]{url} \usepackage{graphicx}   \usepackage{natbib} \usepackage{caption} 

\usepackage{amsmath}
\usepackage{amssymb}
\usepackage{booktabs}
\usepackage{array}
\usepackage{multirow}
\usepackage{tabularx}
\usepackage{float}
\usepackage{placeins}
\usepackage{tcolorbox}
\tcbuselibrary{breakable,skins}
\newtcolorbox{breakablecasebox}{
  enhanced,
  breakable,
  colback=white,
  colframe=black,
  boxrule=0.4pt,
  arc=0pt,
  outer arc=0pt
}

\newcommand{\coderepolink}[1]{\leavevmode\pdfstartlink attr {/Border [0 0 0]} user {/Subtype /Link /A << /S /URI /URI (#1) >>}\url{#1}\pdfendlink}
\newcommand{\citepboxed}[1]{\citep{#1}}

\newcommand{\figrefen}[1]{Figure~\ref{#1}}

\newcommand{\scoreup}[1]{#1}
\newcommand{\scoredown}[1]{#1}
\newcommand{\scorezero}[1]{#1}

\title{\textsc{AMTFV}: Agentic Mathematical Tool-Flow Verification for LLM Self-Correction}
\author{
    Rui Zou\textsuperscript{\rm 1},
    Yutao Zhu\textsuperscript{\rm 1},
    Mengqi Wei\textsuperscript{\rm 2},
    Ji-Rong Wen\textsuperscript{\rm 1}\corresponding
}
\affiliations{
    \textsuperscript{\rm 1}Gaoling School of Artificial Intelligence, Renmin University of China\\
    \textsuperscript{\rm 2}National Engineering Research Center of Educational Big Data, Central China Normal University\\
    \{zouruixyz, ytz, jrwen\}@ruc.edu.cn, weimengqi@mails.ccnu.edu.cn
}

\makeatletter
\g@addto@macro\@maketitle{%
  \g@addto@macro\@thanks{%
    \footnotetext[0]{\textbf{Code}\enspace\coderepolink{https://github.com/TicusFFF/AMTFV}}}}
\makeatother
\begin{document}

\maketitle

\begin{abstract}
Large language models have demonstrated strong mathematical problem-solving capabilities, yet reliably verifying their candidate answers remains challenging. Existing representative methods mainly revise outputs through natural-language reflection or assist verification by directly generating verification programs; the former may not reliably support exact computation, whereas the latter prematurely couples mathematical modeling with low-level implementation. We propose AMTFV (Agentic Mathematical Tool-Flow Verification). By introducing Mathematical Tool Flow (MTF) as an interrupt--execute--resume interface, AMTFV decouples verification modeling from concrete execution and supports exact computation through a mathematical toolbox. Specifically, the verification agent first constructs a verification workflow, encodes the mathematical objects and computational intent requiring reliable execution in an MTF request, and sends it to the mathematical toolbox agent. The latter parses the request, generates executable calls, and dispatches them to the backend for exact computation. Tool outputs then support candidate-answer adjudication, answer revision, and verification-workflow revision. We evaluate AMTFV on five challenging mathematical reasoning datasets with seven model configurations from DeepSeek, GPT, and Gemini. Experimental results show that AMTFV outperforms the representative baselines evaluated in this study overall; under an individual model configuration, it improves average accuracy over the strongest baseline by up to \(8.3\) percentage points, with larger gains on samples of medium and high verification complexity.
\end{abstract}

\section{Introduction}\label{sec:introduction}

Large language models (LLMs) have demonstrated strong mathematical reasoning capabilities~\citepboxed{yang2024qwen25math,deepseekai2025r1,zhan2026mathsmith}. Yet their answers to complex problems may remain unreliable because of computational errors, flawed symbolic derivations, omitted constraints, incomplete enumeration, or incorrect optimality judgments. Prior work further shows that rising answer accuracy can coexist with faulty assumptions, planning failures, and inadequate constraint handling in reasoning chains~\citepboxed{boye2025mathfailures}. A reliable mathematical reasoning system should therefore not only generate answers but also verify that they satisfy the original conditions and revise them when errors are detected~\citepboxed{cobbe2021gsm8k,song-etal-2025-progco}.

Existing backward-verification methods mainly follow two paths. The first revises outputs through natural-language self-reflection, feedback-based rewriting, checklists, or repeated sampling~\citepboxed{pan2024automatically,kamoi2024selfcorrection,madaan2023selfrefine,shinn2023reflexion,cook2024ticking,wang2022selfconsistency}, but does not reliably detect and correct reasoning errors without external feedback~\citepboxed{huang2024selfcorrect,tyen2024llms}. The second augments verification through code execution, such as Python programs~\citepboxed{gao2023pal,chen2023pot,gou2024tora,song-etal-2025-progco}. However, we argue that this can prematurely couple mathematical modeling and verification-target design with low-level implementation. Models are asked to generate executable programs before fully specifying the verification target, forcing them to construct verification objects while handling details such as loop boundaries and numerical precision. Such premature code generation can introduce implementation errors and make verification fragile, with two consequences. First, failures are difficult to localize among mathematical modeling, constraint abstraction, and program boundary handling. Second, exact computation may not be fully delegated to specialized tools, leaving reliability dependent on the model's code-generation ability and ad hoc program quality. Backward verification therefore needs a clearer structure that separates mathematical verification modeling from low-level symbolic compilation, program execution, and exact computation.

This paper proposes \textsc{AMTFV} (Agentic Mathematical Tool-Flow Verification), an autonomous framework for mathematical backward verification and self-correction. At its core is the introduction of Mathematical Tool Flow (MTF) as an intermediate interface that separates mathematical reasoning from concrete execution within the verification process. MTF follows an interrupt--execute--resume interaction pattern: during verification, the LLM emits a local computation request and then pauses, waits for the toolbox to finish execution, and resumes reasoning based on the returned result. In this way, the LLM and computational tools each play to their strengths: the LLM focuses on high-level mathematical reasoning, describing ``what needs to be computed'' solely in terms of mathematical objects and computational intent, and packages this as a structured MTF request. The mathematical toolbox agent receives the request, selects an appropriate mathematical tool according to the computation task (e.g., SymPy~\citepboxed{meurer2017sympy} for symbolic computation and equation solving, or Fraction for exact rational arithmetic), generates an executable call and delegates its execution to the back end, after which the execution result is returned to the verification and correction module for candidate-answer adjudication, answer revision, or verification-workflow revision.

\begin{figure}[t]
\centering
\includegraphics[width=\columnwidth]{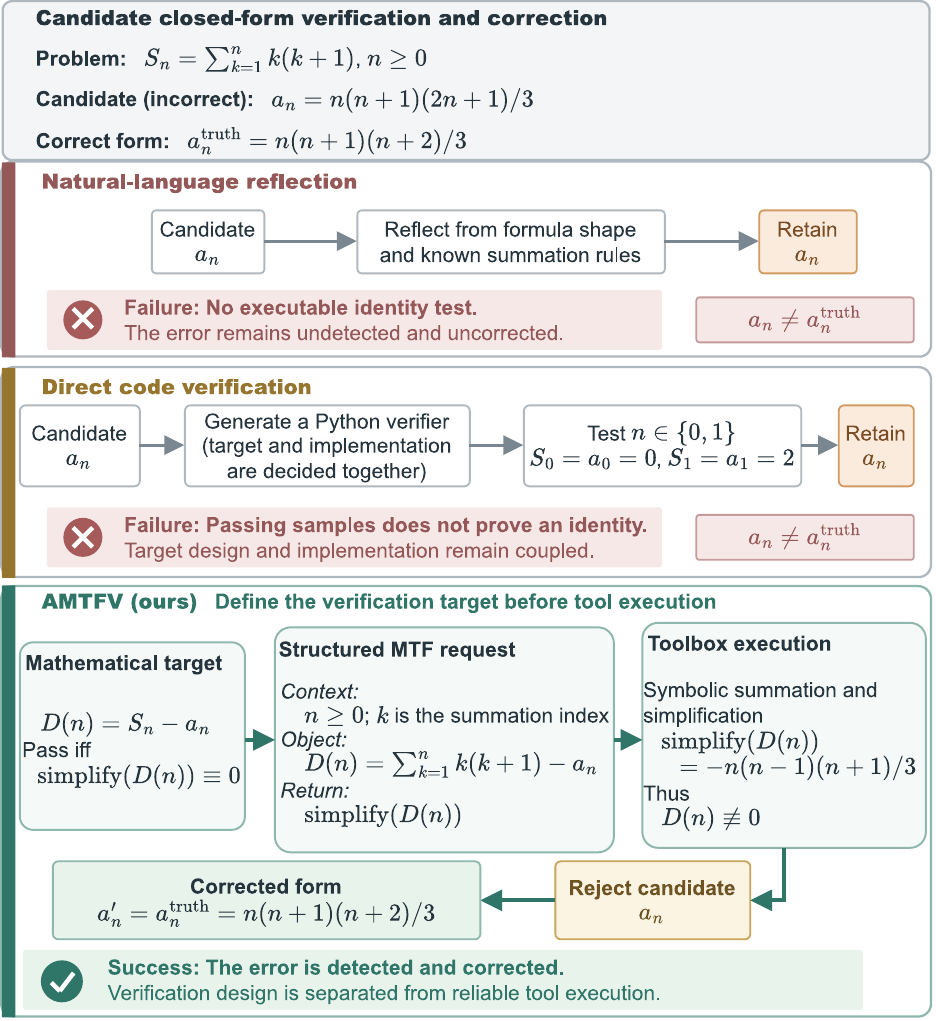}
\caption{Comparison of natural-language reflection, direct code verification, and \textsc{AMTFV} for mathematical answer verification and correction.}
\label{fig:intro_motivation_amtfv}
\end{figure}

This design that decouples reasoning from execution allows the LLM to focus on mathematical modeling without being prematurely drawn into program implementation, delegates formal computation to a tool back-end better suited for precise execution, thereby more fully leveraging the LLM's mathematical reasoning capabilities, and effectively mitigates the computational instability caused by the lack of reliable symbolic support in natural-language reflection and the tight coupling between logic and implementation in ad hoc code-based verification. Moreover, MTF preserves clear mathematical semantics, making the verification intent inspectable, revisable, and reusable. \figrefen{fig:intro_motivation_amtfv} illustrates this distinction with an example of closed-form expression verification and correction: natural-language correction lacks symbolic execution, code-based verification tightly couples the verification target with its implementation, whereas \textsc{AMTFV} first explicitly constructs the verification target and then invokes mathematical tools through MTF, achieving a clean separation between reasoning and execution.

We evaluate \textsc{AMTFV} on diverse mathematical reasoning tasks. In the main DeepSeek experiments, it achieves higher average final accuracy than natural-language reflection, feedback-based rewriting, checklist-guided correction, repeated forward-reasoning sampling, and ProgCo. Supplementary GPT and Gemini experiments likewise show higher average accuracy than verification-enhanced methods such as ProgCo. Compared with the strongest evaluated public baseline, \textsc{AMTFV} improves average accuracy by up to \(8.3\) percentage points. Further analyses suggest more reliable candidate-answer verification and correction, fewer cases where local checks pass despite incorrect final answers, and larger gains on samples of medium and high verification complexity.

Our contributions are threefold: (1) We introduce MTF, an interrupt--execute--resume interface at the core of \textsc{AMTFV}, which decouples mathematical verification modeling from low-level implementation details, avoids premature code generation, and allows LLMs to focus on high-level mathematical reasoning; (2) we introduce a mathematical toolbox agent that translates MTF requests into executable calls for appropriate mathematical tools in the backend, supporting more accurate and comprehensive backward verification of complex mathematical answers; and (3) we validate the effectiveness of \textsc{AMTFV} across diverse mathematical reasoning datasets and multiple mainstream base-model configurations.

\begin{figure*}[t]
    \centering
    \includegraphics[width=1.0\textwidth]{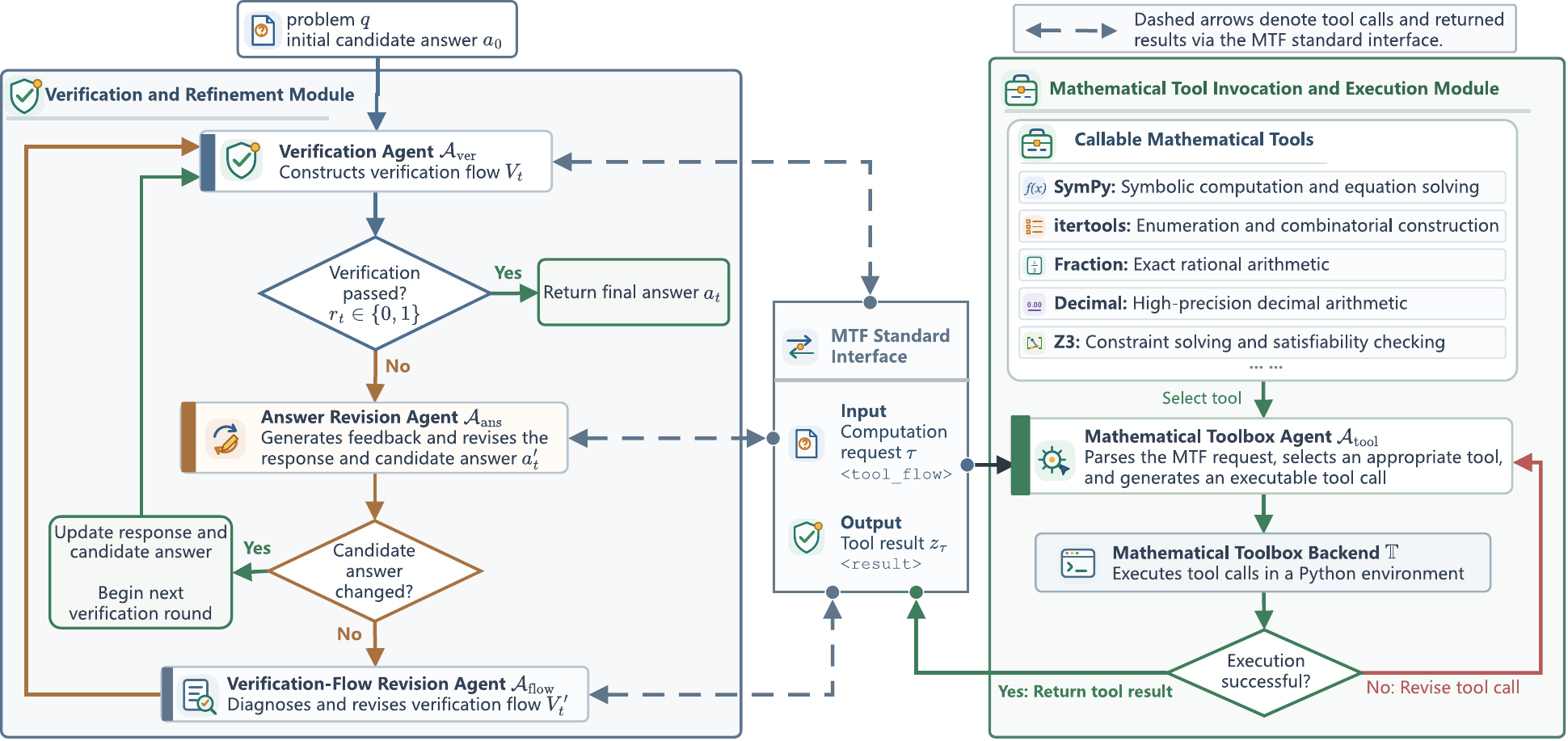}
    \caption{Overview of the \textsc{AMTFV} framework.}
    \label{fig:method_2211}
\end{figure*}

\section{Related Work}\label{sec:related_work}

Our work relates to three research lines: LLM self-correction, tool-augmented mathematical reasoning and agents, and verification-driven reasoning and correction.

\noindent\textbf{LLM self-correction.}
Methods for improving test-time outputs typically use feedback, reflection, checking, or multi-path sampling. Self-Refine iteratively refines outputs with self-generated feedback; Reflexion uses linguistic feedback for subsequent attempts; TICK structures evaluation and improvement with LLM-generated checklists; and Self-Consistency samples multiple reasoning paths and selects a consistent answer for stability~\citepboxed{madaan2023selfrefine,shinn2023reflexion,cook2024ticking,wang2022selfconsistency}. Recent training and inference methods also enhance self-verification and self-correction: S\(^2\)R uses reinforcement learning, while SPOC interleaves solution generation and verification in a single inference pass to trigger spontaneous correction~\citepboxed{ma2025s2r,zhao2025spoc}. Studies show that without reliable external feedback, models do not consistently identify and correct their reasoning errors, especially on complex tasks where revisions may fail or errors are difficult to localize~\citepboxed{pan2024automatically,kamoi2024selfcorrection,huang2024selfcorrect,tyen2024llms}.

\noindent\textbf{Tool-augmented mathematical reasoning and agents.}
Tool-augmented reasoning combines language models with external programs, code interpreters, or specialized tools to mitigate instability in exact computation and symbolic execution. PAL translates mathematical problems into Python-executed programs; Program-of-Thoughts separates numerical computation from natural-language reasoning; and ToRA integrates natural-language reasoning with tool calls for mathematical problem solving~\citepboxed{gao2023pal,chen2023pot,gou2024tora}. Tool-augmented mathematical agents such as AgentMath and R1-Code-Interpreter likewise use code interpreters or tool calls for complex mathematical tasks~\citepboxed{luo2025agentmath,chen2025r1codeinterpreter}. In broader agent research, ReAct interleaves reasoning with external actions, Toolformer learns to invoke APIs, and TRICE uses execution feedback for tool learning, while AutoGen, MetaGPT, and AgentVerse use multi-agent conversations, role specialization, or collaboration for complex tasks~\citepboxed{yao2023react,schick2023toolformer,qiao2024trice,wu2023autogen,hong2023metagpt,chen2023agentverse}.

\noindent\textbf{Verification-driven reasoning and correction.}
Complex mathematical reasoning requires both generating candidate answers and checking them against the original constraints and objective. Early verifier-based work trains verifiers to score or rank candidate solutions and select more reliable answers~\citepboxed{cobbe2021gsm8k}. Recent failure analyses further show that correct final answers need not reflect reliable reasoning: faulty assumptions, planning failures, arithmetic errors, and inadequate constraint handling remain common~\citepboxed{boye2025mathfailures}. Most closely related, ProgCo uses program-driven verification to check candidate answers and program-driven refinement to provide concrete programmatic feedback for self-correction~\citepboxed{song-etal-2025-progco}.

Overall, prior work improves correction through linguistic feedback, external tools, or program-driven verification. In contrast, \textsc{AMTFV} uses MTF as a mathematical-toolbox-oriented intermediate representation to decouple verification modeling from execution and use tool results to guide agentic self-correction, rather than merely adding a code executor.

\section{Method}\label{sec:method}

We develop \textsc{AMTFV}, an agentic mathematical verification and correction framework using MTF as its core interface. Given a problem \(q\) and an initial candidate answer \(a_0\) extracted from the initial response, the system verifies, provides feedback on, and revises the candidate. Whenever verification or revision requires reliable computation, the agents invoke mathematical tools through the standardized MTF interface.

As shown in \figrefen{fig:method_2211}, \textsc{AMTFV} has three components. The left verification and correction module contains a verification agent, an answer-revision agent, and a verification-workflow revision agent. The central standardized MTF interface transmits computation requests and tool results. In the right mathematical tool invocation and execution module, the mathematical toolbox agent \(\mathcal A_{\mathrm{tool}}\) parses MTF requests, selects tools, and generates executable calls, which the mathematical toolbox backend executes. Results return to the left module for adjudication, feedback, and revision. This architecture decouples mathematical verification-target modeling from low-level tool execution. We describe the verification and correction module followed by the mathematical tool invocation and execution module.

\subsection{Verification and Correction Module}\label{sec:mtf_verification_refinement}

Let \(\mathcal A_{\mathrm{ver}}\), \(\mathcal A_{\mathrm{ans}}\), and \(\mathcal A_{\mathrm{flow}}\) denote the verification, answer-revision, and verification-workflow revision agents, respectively. At iteration \(t\), the system first invokes \(\mathcal A_{\mathrm{ver}}\):

\begin{equation}
    \label{eq:loop-verify}
    (V_t,r_t,R_t)=\mathcal A_{\mathrm{ver}}(q,y_t,a_t;V'_{t-1}).
\end{equation}

Here, \(y_t\) is the current response and \(a_t\) its extracted candidate answer. The optional \(V'_{t-1}\) is a reference verification workflow; if unavailable, \(\mathcal A_{\mathrm{ver}}\) reconstructs one from \(a_t\). The executed workflow, adjudication result, and execution record are \(V_t\), \(r_t\in\{0,1\}\), and \(R_t\), respectively. The system returns \(a_t\) if \(r_t=1\); otherwise, it proceeds to feedback and revision.

\paragraph{Candidate Verification.}

At iteration \(t\), \(\mathcal A_{\mathrm{ver}}\) constructs \(V_t\) to determine whether \(a_t\) satisfies the constraints and objective of \(q\). For steps requiring reliable execution, such as symbolic simplification, enumerative counting, exact computation, or constraint solving, it sends MTF requests to \(\mathcal A_{\mathrm{tool}}\). Returned results are written to \(R_t\) and used to produce \(r_t\).
If verification at iteration \(t\) produces an MTF request \(\tau_t^{\mathrm{ver}}\), the invocation is written as
\[
\kappa_t^{\mathrm{ver}}
=
\mathcal A_{\mathrm{tool}}(\tau_t^{\mathrm{ver}};\mathbb T),
\qquad
z_t^{\mathrm{ver}}
=
\operatorname{Run}(\kappa_t^{\mathrm{ver}}).
\]
Here, \(\mathbb T\) is the set of tools exposed by the mathematical toolbox backend, \(\kappa_t^{\mathrm{ver}}\) the generated tool call, and \(z_t^{\mathrm{ver}}\) its result.
\paragraph{Answer Revision.}

When \(r_t=0\), \(a_t\) fails verification and the system invokes \(\mathcal A_{\mathrm{ans}}\). Given the problem \(q\), current response \(y_t\), verification workflow \(V_t\), and execution record \(R_t\), \(\mathcal A_{\mathrm{ans}}\) produces feedback \(F_t\) and a revised response \(y'_t\), from which the system extracts \(a'_t\):
\begin{equation}
\label{eq:loop-revise-answer}
\begin{aligned}
(F_t,y'_t) &= \mathcal A_{\mathrm{ans}}(q,y_t,V_t,R_t),\\
    a'_t &= \operatorname{Extract}(y'_t).
\end{aligned}
\end{equation}
To recompute an expression, enumerate a candidate set, or check constraint feasibility during revision, the agent may produce \(\tau_t^{\mathrm{ans}}\) and invoke the backend through \(\mathcal A_{\mathrm{tool}}\):
\[
    \kappa_t^{\mathrm{ans}}
    =
    \mathcal A_{\mathrm{tool}}(\tau_t^{\mathrm{ans}};\mathbb T),
    \qquad
    z_t^{\mathrm{ans}}
    =
    \operatorname{Run}(\kappa_t^{\mathrm{ans}}).
\]
The result \(z_t^{\mathrm{ans}}\) may be added to \(R_t\) and used to produce \(F_t\) and \(y'_t\). If \(a'_t\neq a_t\), the system updates the response and candidate and verifies the new candidate at the next iteration.

\paragraph{Verification-Workflow Revision.}

If answer revision retains the candidate, i.e., \(a'_t=a_t\), the system invokes \(\mathcal A_{\mathrm{flow}}\). Using the current workflow \(V_t\) and execution record \(R_t\), \(\mathcal A_{\mathrm{flow}}\) diagnoses verification-target coverage, the mathematical objects, and computational evidence, then produces \(V'_t\):
\begin{equation}
    \label{eq:loop-revise-flow}
    V'_t=\mathcal A_{\mathrm{flow}}(q,y_t,V_t,R_t).
\end{equation}
Verification-workflow revision improves checks insufficiently covered by \(V_t\). If the original workflow checks only a subset of candidates, local relations, or intermediate computations, the revision may add the complete set, global optimum, symbolic equivalence, or constraint satisfiability as targets. If diagnosis or revision produces \(\tau_t^{\mathrm{flow}}\), the backend is again invoked through \(\mathcal A_{\mathrm{tool}}\):
\[
\kappa_t^{\mathrm{flow}}
=
\mathcal A_{\mathrm{tool}}(\tau_t^{\mathrm{flow}};\mathbb T),
\qquad
z_t^{\mathrm{flow}}
=
\operatorname{Run}(\kappa_t^{\mathrm{flow}}).
\]
The result \(z_t^{\mathrm{flow}}\) may update \(R_t\) and guide construction of \(V'_t\), which serves as the next iteration's reference workflow.

\paragraph{Iteration Mechanism.}

The response and candidate for the next iteration are updated as follows:
\begin{equation}
\label{eq:loop-state-update}
(y_{t+1},a_{t+1})=
\begin{cases}
(y'_t,a'_t), & a'_t\neq a_t,\\
(y_t,a_t), & a'_t=a_t.
\end{cases}
\end{equation}
If \(a'_t\neq a_t\), the next iteration verifies \(a'_t\); otherwise, it retains \(a_t\) and uses \(V'_t\) as its reference workflow. Iteration stops upon successful verification or at the preset iteration limit.
\subsection{Mathematical Tool Invocation and Execution Module}\label{sec:mtf_interface_execution}

An MTF fragment \(\tau\) is a local mathematical computation request, enclosed by \texttt{<tool\_flow>...</tool\_flow>} tags, that is emitted within the ongoing verification trajectory. When the closing tag is reached, streamed generation is interrupted and the request is sent to the mathematical toolbox. The returned result is appended to the accumulated context, and verification continues in a follow-up model call. A verification trajectory may contain multiple such fragments; their interleaving with reasoning and returned tool results forms the MTF.
Formally,
\[
\tau=\langle \Gamma_\tau,\mathcal M_\tau,\rho_\tau\rangle .
\]
Here, \(\Gamma_\tau\) is the context, such as variables, domains, parameter assumptions, or known constraints; \(\mathcal M_\tau\) is the object to compute, construct, or verify, such as a set, expression, equation system, recurrence, or constraint system; and \(\rho_\tau\) specifies the return operation on \(\mathcal M_\tau\), such as cardinality, symbolic simplification, feasible solutions, a global optimum, or satisfiability.

The tuple schema for \(\tau\) can represent different verification tasks. The following example maps a symbolic-expression verification target to an MTF request and tool call.

\begin{tcolorbox}[
    enhanced,
    breakable,
    width=0.97\columnwidth,
    colback=white,
    colframe=black,
    boxrule=\fboxrule,
    arc=0pt,
    outer arc=0pt,
    boxsep=0pt,
    left=\fboxsep,
    right=\fboxsep,
    top=\fboxsep,
    bottom=\fboxsep
]
\textbf{Example: Symbolic-expression verification.}
    A chocolate bar costs \(c\) and a vanilla bar \(c+2\). Jamie buys one chocolate and three vanilla bars, while Kevin buys five chocolate bars. The candidate answer is \(8c+6\).
\par\noindent
Let \(\Delta(c)=c+3(c+2)+5c-(8c+6)\). The candidate passes verification if and only if \(\operatorname{Simplify}(\Delta(c))\equiv0\).
\par\noindent
\textbf{MTF request.}\par\noindent
\begin{tabular}{@{\hspace{1.5em}}l@{}}
\texttt{<tool\_flow>}\\
\texttt{Context: c is symbolic.}\\
\texttt{Object: Delta(c)=}\\
\quad\texttt{c+3*(c+2)+5*c-(8*c+6).}\\
\texttt{Return: simplify(Delta(c)).}\\
\texttt{</tool\_flow>}
\end{tabular}
\par\noindent
{\raggedright This corresponds to \(\tau=\langle\{c\text{ is symbolic}\},\Delta(c),\operatorname{Simplify}\rangle\).\par}
\par\noindent
\textbf{Tool call and result.}\par\noindent
\begin{tabular}{@{\hspace{1.5em}}l@{}}
\texttt{c = sympy.symbols("c")}\\
\texttt{D = c+3*(c+2)+5*c-(8*c+6)}\\
\texttt{z\_tau = sympy.simplify(D)}
\end{tabular}
\par\noindent
The backend returns \(z_\tau=c\not\equiv0\); hence the candidate fails verification and \(r_t=0\).
\end{tcolorbox}

Here, \(\rho_\tau=\operatorname{Simplify}\) specifies symbolic verification. Counting, optimization, and constraint solving use the same template, replacing \(\mathcal M_\tau\) with a set, objective, or constraint system and \(\rho_\tau\) with cardinality, optimum, satisfiability, or solution set. MTF thus uniformly expresses ``context--object--return specification'' for different verification targets.
MTF represents mathematical computational intent, and \(\mathcal A_{\mathrm{tool}}\) translates it into an executable call. Through \(\tau\), the three agents specify the object and desired return; \(\mathcal A_{\mathrm{tool}}\) selects a tool, generates a call, and dispatches it to the backend. Together, MTF and \(\mathcal A_{\mathrm{tool}}\) form an intermediate computational interface to the toolbox.
MTF also makes complete verification objects explicit. For counting, optimization, symbolic verification, or constraint solving, the backend may return a full set and its cardinality, a global optimum, a symbolic difference, or satisfiability. This helps \textsc{AMTFV} avoid checking only local candidate consistency while overlooking the complete target.

Let the set of tools exposed by the mathematical toolbox backend be
\[
\mathbb T=
\{
\mathcal T_{\mathrm{sym}},
\mathcal T_{\mathrm{enum}},
\mathcal T_{\mathrm{exact}},
\mathcal T_{\mathrm{smt}},
\ldots
\}.
\]
where \(\mathcal T_{\mathrm{sym}}\), \(\mathcal T_{\mathrm{enum}}\), \(\mathcal T_{\mathrm{exact}}\), and \(\mathcal T_{\mathrm{smt}}\) support symbolic computation, enumeration, exact numerical computation, and constraint solving, respectively.

Given \(\tau\) and \(\mathbb T\), \(\mathcal A_{\mathrm{tool}}\) generates an executable call \(\kappa_\tau\):
\[
\kappa_\tau
=
\mathcal A_{\mathrm{tool}}(\tau;\mathbb T)
=
(\mathcal T_j,u_j),
\qquad
\mathcal T_j\in\mathbb T.
\]
Here, \(\mathcal T_j\) is the selected tool and \(u_j\) its input. The agent selects \(\mathcal T_j\) according to \(\rho_\tau\) and constructs \(u_j\) from \(\Gamma_\tau\) and \(\mathcal M_\tau\).

The backend executes the call and returns
\[
z_\tau
=
\operatorname{Run}(\kappa_\tau)
=
\mathcal T_j(u_j).
\]
Here, \(z_\tau\) is the result specified by \(\rho_\tau\).

The result \(z_\tau\) is enclosed in \texttt{<result>} and appended to execution record \(R\):
\[
R
\leftarrow
\operatorname{Append}
\bigl(R,(\tau,\mathcal T_j,u_j,z_\tau)\bigr).
\]
The three agents use results in \(R\) for candidate-answer adjudication, answer revision, or workflow revision.

Our mathematical toolbox backend runs in Python: SymPy supports symbolic computation and equation solving, itertools combinatorial enumeration, and Fraction exact rational arithmetic. Python serves only as the environment for calls generated by \(\mathcal A_{\mathrm{tool}}\).

\begin{table*}[t]
\centering
\small
\setlength{\tabcolsep}{1.75pt}
\begin{tabular}{@{}lcccccc@{\hspace{11pt}}cccccc@{\hspace{11pt}}cccccc@{}}
\toprule
& \multicolumn{6}{c}{DeepSeek-Flash}
& \multicolumn{6}{c}{DeepSeek-Flash-Think}
& \multicolumn{6}{c}{DeepSeek-Pro} \\
\cmidrule(lr){2-7}\cmidrule(lr){8-13}\cmidrule(lr){14-19}
Method
& A24 & A25 & B25 & HMM & AMO & Avg
& A24 & A25 & B25 & HMM & AMO & Avg
& A24 & A25 & B25 & HMM & AMO & Avg \\
\midrule
Initial Score
& \(63.3\) & \(43.3\) & \(86.7\) & \(33.3\) & \(14.0\) & \(44.1\)
& \(100.0\) & \(100.0\) & \(100.0\) & \(100.0\) & \(62.0\) & \(88.8\)
& \(63.3\) & \(53.3\) & \(60.0\) & \(26.7\) & \(10.0\) & \(38.8\) \\
reflex
& \(63.3\) & \(46.7\) & \(83.3\) & \(33.3\) & \(18.0\) & \(45.3\)
& \(96.7\) & \(100.0\) & \(93.3\) & \(93.3\) & \(56.0\) & \(84.1\)
& \(60.0\) & \(63.3\) & \(70.0\) & \(40.0\) & \(10.0\) & \(44.1\) \\
Self-Refine
& \(66.7\) & \(60.0\) & \(86.7\) & \(50.0\) & \(14.0\) & \(50.6\)
& \(93.3\) & \(100.0\) & \(90.0\) & \(96.7\) & \(56.0\) & \(83.5\)
& \(66.7\) & \(66.7\) & \(86.7\) & \(56.7\) & \(16.0\) & \(53.5\) \\
Self-Refl.
& \(63.3\) & \(46.7\) & \(86.7\) & \(33.3\) & \(16.0\) & \(45.3\)
& \(96.7\) & \(100.0\) & \(93.3\) & \(100.0\) & \(60.0\) & \(86.5\)
& \(66.7\) & \(60.0\) & \(73.3\) & \(40.0\) & \(8.0\) & \(44.7\) \\
CheckList
& \(60.0\) & \(56.7\) & \(80.0\) & \(30.0\) & \(16.0\) & \(44.7\)
& \(96.7\) & \(100.0\) & \(86.7\) & \(86.7\) & \(56.0\) & \(81.8\)
& \(66.7\) & \(53.3\) & \(70.0\) & \(40.0\) & \(16.0\) & \(45.3\) \\
CoT-Tool
& \(76.7\) & \(60.0\) & \(73.3\) & \(43.3\) & \(4.0\) & \(45.9\)
& \(93.3\) & \(93.3\) & \(93.3\) & \(86.7\) & \(44.0\) & \(77.6\)
& \(56.7\) & \(43.3\) & \(43.3\) & \(23.3\) & \(6.0\) & \(31.2\) \\
ProgCo
& \(66.7\) & \(53.3\) & \(96.7\) & \(43.3\) & \(18.0\) & \(51.2\)
& \(100.0\) & \(100.0\) & \(100.0\) & \(100.0\) & \(58.0\) & \(87.6\)
& \(66.7\) & \(60.0\) & \(66.7\) & \(40.0\) & \(12.0\) & \(44.7\) \\
ProgCo-Py
& \(66.7\) & \(60.0\) & \(90.0\) & \(43.3\) & \(24.0\) & \(52.9\)
& \(100.0\) & \(100.0\) & \(100.0\) & \(100.0\) & \(60.0\) & \(88.2\)
& \(80.0\) & \(60.0\) & \(73.3\) & \(43.3\) & \(14.0\) & \(49.4\) \\
\textbf{\textsc{AMTFV}}
& \(\mathbf{80.0}\) & \(\mathbf{66.7}\) & \(\mathit{93.3}\) & \(\mathbf{60.0}\) & \(\mathbf{28.0}\) & \(\mathbf{61.2}\)
& \(\mathbf{100.0}\) & \(\mathbf{100.0}\) & \(\mathbf{100.0}\) & \(\mathbf{100.0}\) & \(\mathbf{64.0}\) & \(\mathbf{89.4}\)
& \(\mathit{76.7}\) & \(\mathbf{66.7}\) & \(\mathbf{93.3}\) & \(\mathit{53.3}\) & \(\mathbf{20.0}\) & \(\mathbf{57.1}\) \\
\bottomrule
\end{tabular}
\caption{Main results. Avg denotes sample-weighted average accuracy across the five datasets. In the \textsc{AMTFV} row, bold indicates the highest or tied-highest value among all correction methods in the corresponding column, and italics indicate the second-highest value.}
\label{tab:main_results}
\end{table*}

\begin{figure*}[t]
\centering
\includegraphics[width=0.98\textwidth]{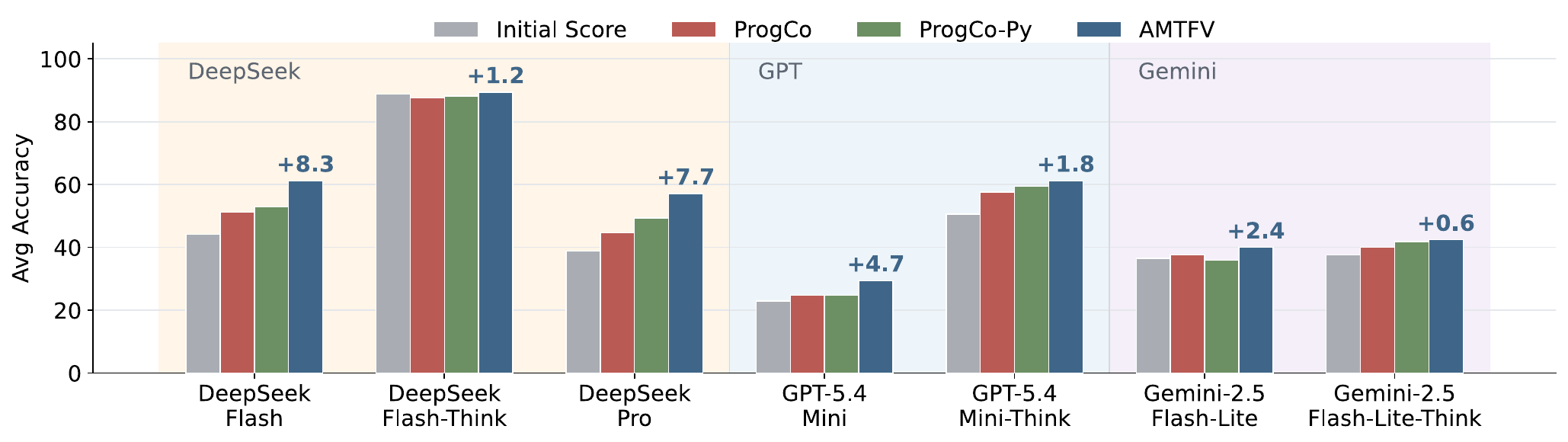}
\caption{Average accuracy of verification-enhanced methods under each base-model setting. The horizontal axis shows the seven base-model settings, and the vertical axis shows sample-weighted average accuracy (\%) across the five datasets. Gray bars denote Initial Score, while the other bars show final accuracy with \(\texttt{max\_turn}=3\). Blue annotations report the absolute improvement of \textsc{AMTFV} over the better result between ProgCo and ProgCo-Py, in percentage points.}
\label{fig:architecture_variants}
\end{figure*}

\section{Experiments}\label{sec:experiment}

We evaluate \textsc{AMTFV} for mathematical reasoning verification and correction against self-correction, reasoning-enhanced, and verification-enhanced methods. We analyze its performance and gains through cross-model architectural comparisons, correction-state transitions, and verification-complexity bins. The supplementary material covers iteration budgets, correlations with empirical difficulty, MTF call types (see Supplementary Figure~\ref{fig:app_mtf_tool_types}), and case processes.

\subsection{Experimental Setup}\label{sec:experiment_setup}

\textbf{Datasets.}
We use five hard-test mathematical reasoning datasets totaling 170 problems, abbreviated as A24, A25, B25, HMM, and AMO. \textbf{AIME 2024}~\citepboxed{huggingfaceh42025aime2024} and \textbf{AIME 2025}~\citepboxed{yentinglin2025aime} each contain 30 American Invitational Mathematics Examination problems assessing multistep reasoning and exact computation. \textbf{BRUMO 2025} contains 30 challenging problems assessing complex-condition understanding and multistep reasoning, and \textbf{HMMT February 2025} contains 30 competition problems spanning algebra, geometry, combinatorics, and number theory~\citepboxed{balunovic2025matharena}. \textbf{AMO Bench} contains 50 Olympiad-level problems with more diverse problem and answer formats~\citepboxed{an2025amobench}.

\textbf{Base models.}
We cover DeepSeek, GPT, and Gemini. The main experiments use \textbf{DeepSeek-Flash}, \textbf{DeepSeek-Flash-Think}, and \textbf{DeepSeek-Pro} to compare correction across non-thinking, thinking, and stronger-model settings~\citepboxed{deepseek2026v4preview}. For cross-model generalization, we further compare ProgCo, ProgCo-Py, and \textsc{AMTFV} using \textbf{GPT-5.4-Mini} and \textbf{GPT-5.4-Mini-Think}~\citepboxed{openai2026models}, and \textbf{Gemini-2.5-Flash-Lite} and \textbf{Gemini-2.5-Flash-Lite-Think}~\citepboxed{geminiModelsDocs,geminiOpenAICompat}.

\textbf{Compared methods.}
The main experiments compare three method classes. Natural-language correction includes \textbf{reflex}, the vanilla-reflex configuration from ProgCo; \textbf{Self-Refine}~\citepboxed{madaan2023selfrefine}; \textbf{Self-Reflection}~\citepboxed{shinn2023reflexion}, abbreviated as \textbf{Self-Refl.} in the table; and \textbf{CheckList}~\citepboxed{cook2024ticking}. They represent vanilla reflection, iterative self-feedback, self-reflective feedback, and checklist-guided correction, respectively. The reasoning-enhanced \textbf{CoT-Tool}~\citepboxed{abedi_cogitator_2025,wei2022chain,wang2022selfconsistency} combines the Cogitator toolkit, multi-path chain-of-thought reasoning, and Self-Consistency selection. Verification-enhanced methods include \textbf{ProgCo}, which uses program-driven verification and refinement, and \textbf{ProgCo-Py}, our setting that enables Python-tool feedback in the official implementation~\citepboxed{song-etal-2025-progco}. Cross-model and progressive comparisons focus on ProgCo, ProgCo-Py, and \textsc{AMTFV}, tracing the progression from programmatic verification through general Python-tool feedback to an MTF-based framework that decouples mathematical tool flows.

\textbf{Evaluation settings and metrics.}
For each problem, the base model produces an initial response from which we extract a candidate answer. Except for Initial Score, all methods receive the same initial response and candidate. We set \(\texttt{max\_turn}=3\) for methods controlling iterative correction or tool-use rounds; others follow their standard procedures. The primary metric is final-answer accuracy, with cross-dataset Avg weighted by dataset size. The supplementary material details implementation and fairness settings (see Supplementary Tables~\ref{tab:app_model_configurations} and~\ref{tab:app_method_budgets}). We also analyze correction-state transitions and verification-complexity bins.

\subsection{Main Results}\label{sec:main_results}

Table~\ref{tab:main_results} compares the final-answer accuracy of \textsc{AMTFV} and representative baselines under the three main evaluation settings: DeepSeek-Flash, DeepSeek-Flash-Think, and DeepSeek-Pro.
\textsc{AMTFV} achieves the highest average accuracy under all three DeepSeek settings. From Initial Score, it improves DeepSeek-Flash from 44.1 to 61.2, DeepSeek-Flash-Think from 88.8 to 89.4, and DeepSeek-Pro from 38.8 to 57.1: gains of \(+17.1\), \(+0.6\), and \(+18.3\) percentage points, respectively. With identical initial answers and a fixed maximum iteration budget, these results suggest that MTF-based verification and correction can improve final-answer accuracy more than the compared methods under these settings.

The smaller DeepSeek-Flash-Think gain mainly reflects Initial Scores of 100.0 on A24, A25, B25, and HMM, leaving most room for correction on AMO. Most methods decline from this strong initial setting, indicating overcorrection risk. In contrast, \textsc{AMTFV} preserves accuracy on the four saturated datasets, improves AMO from 62.0 to 64.0, and raises average accuracy from 88.8 to 89.4, indicating that it better avoids aggregate degradation.

\textsc{AMTFV} also exceeds ProgCo and ProgCo-Py in average accuracy under all three DeepSeek settings. Over the better ProgCo-family result, its gains are approximately \(+8.3\), \(+1.2\), and \(+7.7\) percentage points on DeepSeek-Flash, DeepSeek-Flash-Think, and DeepSeek-Pro. We next compare their progressive architectural differences under additional base models.

\subsection{Architectural Variant Analysis}\label{sec:architecture_variants}

We further compare ProgCo, ProgCo-Py, and \textsc{AMTFV} across base models. They form a progressive architectural sequence: ProgCo uses program-driven verification and correction; ProgCo-Py adds Python execution feedback; and \textsc{AMTFV} adds MTF, a mathematical toolbox interface, and closed-loop verification-workflow revision. Unless stated otherwise, \(\texttt{max\_turn}=3\).

Figure~\ref{fig:architecture_variants} compares average accuracy across all seven base-model settings. \textsc{AMTFV} ranks highest in every setting, exceeding the better of ProgCo and ProgCo-Py by \(+0.6\) to \(+8.3\) percentage points; its gains thus extend beyond DeepSeek. The supplementary material reports per-dataset GPT and Gemini results (see Supplementary Tables~\ref{tab:appendix_gpt_progco_series_combined} and~\ref{tab:appendix_gemini_progco_series_combined}).
Python execution feedback lets ProgCo-Py improve over ProgCo in some settings, but inconsistently. By contrast, \textsc{AMTFV} uses MTF to specify the mathematical objects, constraints, and computational objective before backend execution, then applies tool results to adjudication, answer revision, and verification-workflow revision. The comparison supports our design motivation: under the evaluated settings, decoupling verification targets from low-level execution and incorporating tool results into a multistep verification--correction loop can be more stable than adding general program-execution feedback alone.

\begin{figure}[!htbp]
\centering
\includegraphics[width=0.95\linewidth]{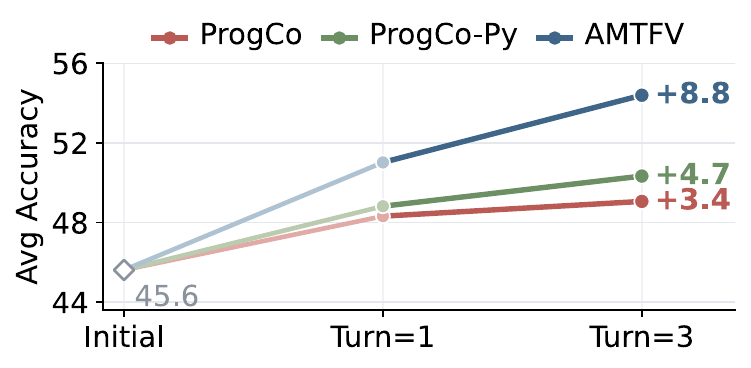}
\caption{Overall average-accuracy trajectories under different iteration budgets. The horizontal axis shows Initial Score, \(\texttt{max\_turn}=1\), and \(\texttt{max\_turn}=3\); the vertical axis shows average accuracy across seven base-model settings. Curve-end values denote absolute improvements from Initial Score to \(\texttt{max\_turn}=3\), in percentage points.}
\label{fig:iteration_overall_trend}
\end{figure}

Figure~\ref{fig:iteration_overall_trend} suggests that, under a fixed budget, \textsc{AMTFV} uses additional verification--correction rounds more effectively. The supplementary material provides trajectories for individual model settings (see Supplementary Figure~\ref{fig:app_iteration_budget}).

\subsection{Correction Behavior and Verification Complexity Analysis}\label{sec:correction_complexity}

We analyze \textsc{AMTFV}'s gains through correction-state transitions, which measure error correction and preservation of correct answers, and verification-complexity bins, which show whether gains concentrate on examples requiring more complex interactions.

\subsubsection{Correction Behavior Analysis}

We characterize correction behavior with row-normalized transition matrices, where \(C\) and \(W\) denote correct and incorrect answers. \(W\rightarrow C\) measures error correction, whereas \(C\rightarrow W\) measures the risk of corrupting a correct answer.

\begin{figure}[t]
\centering
\includegraphics[width=0.98\linewidth]{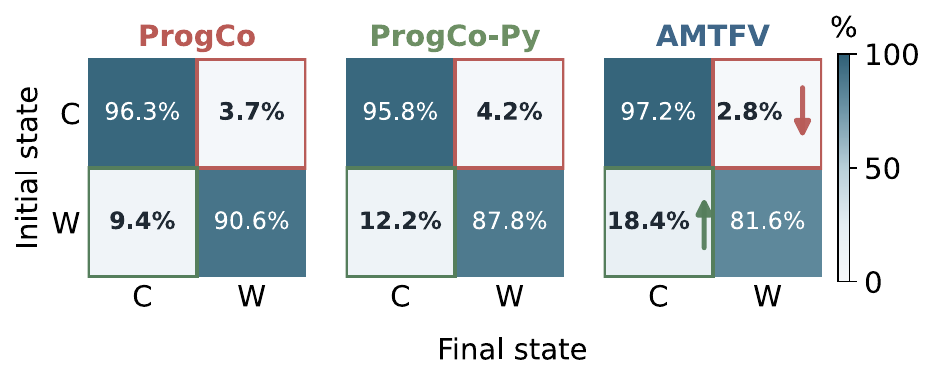}
\caption{Row-normalized correction-state transition matrices. Rows indicate initial correctness and columns indicate final correctness, where \(C\) denotes a correct answer and \(W\) an incorrect answer. Each cell reports the percentage of examples in the corresponding initial state that transition to the given final state, with each row normalized to 100\%. Green borders and arrows highlight error-correcting transitions \(W\rightarrow C\), while red borders and arrows highlight error-introducing transitions \(C\rightarrow W\).}
\label{fig:correction_transition}
\end{figure}

Figure~\ref{fig:correction_transition} aggregates \(7\times170=1190\) model--problem instances across seven base models. \textsc{AMTFV}'s \(W\rightarrow C\) rate is 18.4\%, versus 9.4\% for ProgCo and 12.2\% for ProgCo-Py. Its \(C\rightarrow W\) rate is 2.8\%, below ProgCo's 3.7\% and ProgCo-Py's 4.2\%. Thus, its gains primarily reflect better error correction and preservation of correct answers rather than aggressive rewriting.

\subsubsection{Verification Complexity Analysis}

\begin{figure}[t]
\centering
\includegraphics[width=0.98\linewidth]{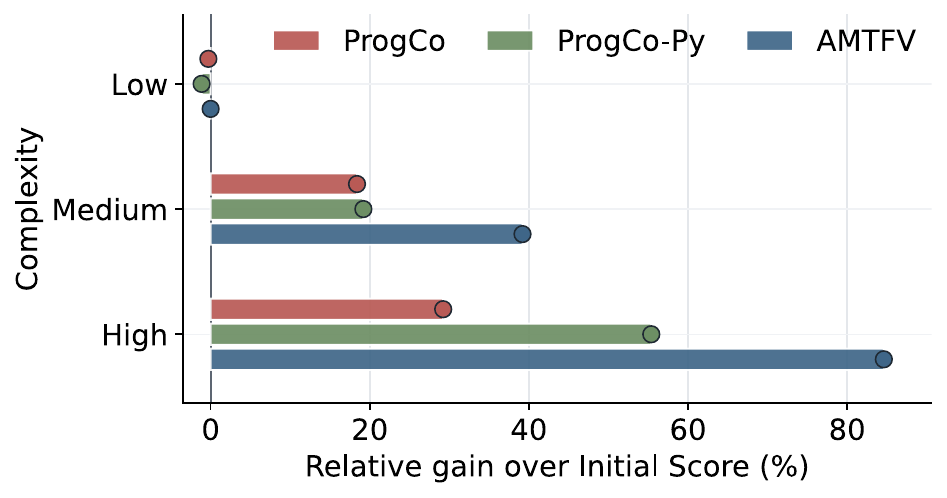}
\caption{Normalized accuracy gains over Initial Score across verification-complexity bins. The horizontal axis reports \((\mathrm{Final\ Score}-\mathrm{Initial\ Score})/\mathrm{Initial\ Score}\), the accuracy gain normalized by initial accuracy. Low, Medium, and High bins are defined in the supplementary material.}
\label{fig:complexity_bin_accuracy}
\end{figure}

We use the MTF call count per model--problem instance as an operational measure of verification complexity because it reflects the amount of computational interaction required for verification. The Low, Medium, and High bins correspond to at most one, two to three, and at least four MTF calls, respectively. The analysis covers \(7\times170=1190\) instances and compares all methods on the same examples per bin. Supplementary Figure~\ref{fig:app_difficulty_complexity} shows a positive relationship between problem-level verification complexity and empirical difficulty: Pearson \(r=0.67\) and Spearman \(\rho=0.70\).

Figure~\ref{fig:complexity_bin_accuracy} shows little gain from any verification-enhanced method in the Low bin. Normalized by Initial Score, \textsc{AMTFV}'s gains are \(39.2\%\) and \(84.6\%\) in the Medium and High bins, exceeding ProgCo-Py by approximately \(20.0\) and \(29.2\) percentage points. Its advantage is therefore strongest when more verification interactions are required. The supplementary material provides the empirical-difficulty--verification-complexity analysis and complexity definition.

The supplementary material presents successful cases in minimality verification, combinatorial counting, lattice-point enumeration, and exact double summation (see Supplementary Table~\ref{tab:successful_cases}). They show how \textsc{AMTFV} uses MTF results for global constraints, exhaustive enumeration, and exact-computation verification, and detail an exact double-summation case from the B25 dataset.

\section{Conclusion}\label{sec:conclusion}

We introduce \textsc{AMTFV} to decouple verification-target design from low-level implementation in mathematical backward verification. MTF separates mathematical verification modeling from tool execution, and its returned results support candidate adjudication and self-correction. Experiments show that \textsc{AMTFV} outperforms representative correction and verification methods across mathematical reasoning datasets and base models, with larger gains at medium or high verification complexity. Further correction-state analysis indicates that these gains primarily arise from correcting initially incorrect answers while preserving initially correct ones, rather than from aggressive rewriting. Future work will extend the framework beyond mathematical toolboxes to domain-specific systems, applying the ``task modeling--specialized tool execution'' paradigm to scientific tasks requiring reliable verification and computation.

\bibliography{amtfv}

\clearpage
\onecolumn
\raggedbottom
\renewcommand{\scoreup}[1]{\textcolor{green!50!black}{#1}}
\renewcommand{\scoredown}[1]{\textcolor{red}{#1}}
\renewcommand{\scorezero}[1]{\textcolor{black}{#1}}
\appendix
\setcounter{secnumdepth}{2}
\begin{center}
{\LARGE\bfseries Supplementary Material}
\end{center}

The main text reports complete results under the primary DeepSeek evaluation settings. This supplementary document provides implementation details and fair-comparison settings, together with detailed results for verification-enhanced methods under GPT and Gemini, statistical significance and computational cost analyses, iteration-budget analyses, the relationship between empirical difficulty and verification complexity, MTF call types, successful cases, and complete case-process comparisons. These materials further document \textsc{AMTFV}'s evaluation protocol, cross-model performance, runtime behavior, verification-complexity signal, distribution of tool calls, and verification--correction process.

\section{Implementation Details and Fair-Comparison Protocol}\label{app:implementation_details}

This section describes base-model configurations, shared initialization, generation parameters, method budgets, and answer evaluation. Experimental settings are controlled separately for each base model. For example, under DeepSeek-Flash, all compared methods use the same model identifier and reasoning mode and, when applicable, share the same initial response, candidate answer, temperature, maximum number of outer iterations, and final evaluation criteria. Each GPT and Gemini setting follows the same principle. Here, ``the same configuration'' refers only to comparisons among methods using the same base model; it does not imply identical configurations across different base models.

\subsection{Base Models and Reasoning Configurations}

Table~\ref{tab:app_model_configurations} lists the model settings used in this paper and their corresponding official model identifiers. Names with the \texttt{-Think} suffix are experimental labels used to distinguish reasoning configurations, not additional official model identifiers. For example, GPT-5.4-Mini and GPT-5.4-Mini-Think use the same base model with different reasoning configurations; the two Gemini settings likewise use the same base model and differ only in whether thinking is enabled. Model identifiers and descriptions of reasoning capabilities follow the corresponding official documentation~\citepboxed{deepseek2026v4preview,openai2026models,geminiModelsDocs,geminiOpenAICompat}.

\begin{table}[H]
\centering
\small
\setlength{\tabcolsep}{5pt}
\renewcommand{\arraystretch}{1.08}
\begin{tabularx}{0.92\textwidth}{l l X}
\hline
Model setting & Official model identifier & Reasoning configuration \\
\hline
DeepSeek-Flash & \texttt{deepseek-v4-flash} & Non-Thinking \\
DeepSeek-Flash-Think & \texttt{deepseek-v4-flash} & Thinking enabled, reasoning effort: high \\
DeepSeek-Pro & \texttt{deepseek-v4-pro} & Non-Thinking \\
GPT-5.4-Mini & \texttt{gpt-5.4-mini} & Reasoning effort: none \\
GPT-5.4-Mini-Think & \texttt{gpt-5.4-mini} & Reasoning effort: low \\
Gemini-2.5-Flash-Lite & \texttt{gemini-2.5-flash-lite} & Thinking disabled \\
Gemini-2.5-Flash-Lite-Think & \texttt{gemini-2.5-flash-lite} & Thinking enabled \\
\hline
\end{tabularx}
\caption{Base models and reasoning configurations. \texttt{-Think} denotes an experimental setting in which the corresponding reasoning mode is enabled.}
\label{tab:app_model_configurations}
\end{table}

\subsection{Shared Initialization and Within-Model Fairness}

For each base-model--problem pair, we first use the base model to generate an initial response and extract its initial candidate answer. reflex, Self-Refine, Self-Refl., CheckList, ProgCo, ProgCo-Py, and \textsc{AMTFV} all begin from the same corresponding pregenerated initial response and candidate answer. During execution, these shared initial outputs are loaded in read-only form rather than resampled by each method. This design reduces variation from initial responses, so observed differences under the same base-model setting more directly reflect the methods' reflection, verification, tool invocation, or correction procedures. Because CoT-Tool does not revise an existing response, it directly generates candidate paths from the original problem according to its standard multi-path reasoning procedure.

Each method retains the prompt templates and procedures specified by its original paper or official implementation. For a given base model, all methods use the same model identifier, reasoning mode, and applicable shared inputs. Different base models retain their respective official default configurations; to preserve their individual characteristics, we do not force GPT, Gemini, and DeepSeek to use the same absolute context length.

\subsection{Generation Parameters and Method Budgets}

Model requests use a common OpenAI-compatible Chat Completions format. Whenever the parameter is supported, temperature is set to \(0\). The context window, maximum output length, and other unspecified sampling parameters retain the defaults of the corresponding model API. Thus, all methods using the same base model share that API's default context and output limits, while different base models retain their own defaults.

The iteration-budget analysis evaluates \(\texttt{max\_turn}\in\{1,3\}\), while the main experiments use \(\texttt{max\_turn}=3\), giving all iterative methods the same upper bound on outer correction rounds. The remaining parameters follow the original method specifications or the defaults of the corresponding model API.

For methods with iterative correction or verification structures, the maximum number of outer iterations is uniformly set to \(\texttt{max\_turn}=3\). Following its multi-path consistency procedure, CoT-Tool independently generates and selects among \(3\) reasoning paths. Table~\ref{tab:app_method_budgets} summarizes the outer reasoning budget for each method. Thus, every iterative method runs for at most \(3\) rounds; because their internal structures differ, the numbers of model calls, feedback steps, or tool calls within each round are not forced to be identical.

Together with the shared initialization described above, the official pipeline performs no randomized local sampling, data shuffling, or stochastic preprocessing, so no local random seed is required. Each reported model--method--problem result comes from one official run, and the supplementary material includes the raw outputs used to compute the reported aggregates.

\begin{table}[H]
\centering
\small
\setlength{\tabcolsep}{4pt}
\renewcommand{\arraystretch}{1.08}
\begin{tabularx}{\linewidth}{>{\raggedright\arraybackslash}X l}
\hline
Method & Outer reasoning budget \\
\hline
Initial Score & No additional correction \\
reflex / Self-Refine / Self-Refl. / CheckList & At most \(3\) correction rounds \\
ProgCo / ProgCo-Py / \textsc{AMTFV} & At most \(3\) verification--correction rounds \\
CoT-Tool & \(3\) independent reasoning paths \\
\hline
\end{tabularx}
\caption{Outer reasoning budgets used by different methods.}
\label{tab:app_method_budgets}
\end{table}

\subsection{Mathematical Tool Execution}

Both ProgCo-Py and \textsc{AMTFV} can receive Python execution feedback, but each organizes tool use according to its own method definition. \textsc{AMTFV} uses MTF to express the mathematical object to be verified and the desired return, after which the mathematical toolbox agent selects and invokes tools such as SymPy, itertools, Fraction, Decimal, or Z3. Tool requests, generated calls, returned results, and execution errors are retained to support subsequent adjudication and correction. Different problems and model settings may run in parallel.

\subsection{Answer Extraction and Unified Evaluation}

Each method obtains a candidate answer from its final response using its predefined output format and extraction procedure. Once extracted, every candidate is compared with the gold answer through the same three-stage equivalence evaluation. The first stage performs normalized exact-string matching. The second applies floating-point comparison to numerically parseable answers with a tolerance of \(10^{-3}\). If the first two stages are inconclusive, the third uses the same DeepSeek-Flash evaluator at temperature \(0\) to determine whether the two final answers are mathematically equivalent. For set-valued, multi-solution, and sequence answers, the evaluator also checks element completeness and ordering requirements; incomplete solution sets are not considered equivalent. All methods use the same gold answers and evaluation rules. Final accuracy is the proportion of correctly answered examples, and the cross-dataset Avg is weighted by dataset size.

\paragraph{Code Availability.}
Code is available at \coderepolink{https://github.com/TicusFFF/AMTFV}.

\section{Detailed Results and Computational Cost}

\subsection{Supplementary Results on GPT and Gemini}\label{app:gpt_gemini_supplementary_results}

This section reports detailed average accuracies for ProgCo, ProgCo-Py, and \textsc{AMTFV} under the GPT and Gemini settings, supporting the progressive architectural comparison in the main text. Supplementary Tables~\ref{tab:appendix_gpt_progco_series_combined} and~\ref{tab:appendix_gemini_progco_series_combined} report results for both \(\texttt{max\_turn}=1\) and \(\texttt{max\_turn}=3\) to show how each method behaves as the fixed iteration budget changes.

\begin{table}[H]
\centering
\small
\setlength{\tabcolsep}{3pt}
\renewcommand{\arraystretch}{1.08}
\begin{tabular}{llcccccccccccc}
\hline
Turn & Method & \multicolumn{6}{c}{GPT-5.4-Mini} & \multicolumn{6}{c}{GPT-5.4-Mini-Think} \\
\cline{3-14}
& & A24 & A25 & B25 & HMM & AMO & Avg & A24 & A25 & B25 & HMM & AMO & Avg \\
\hline
\multirow{3}{*}{1} & ProgCo & \(40.0\) & \(26.7\) & \(43.3\) & \(13.3\) & \(8.0\) & \(24.1_{\scoreup{+1.2}}\) & \(70.0\) & \(80.0\) & \(76.7\) & \(53.3\) & \(22.0\) & \(55.9_{\scoreup{+5.3}}\) \\
 & ProgCo-Py & \(46.7\) & \(26.7\) & \(43.3\) & \(13.3\) & \(4.0\) & \(24.1_{\scoreup{+1.2}}\) & \(73.3\) & \(73.3\) & \(80.0\) & \(56.7\) & \(26.0\) & \(57.6_{\scoreup{+7.0}}\) \\
 & \textsc{AMTFV} & \(\mathbf{50.0}\) & \(\mathbf{43.3}\) & \(\mathbf{53.3}\) & \(\mathbf{16.7}\) & \(\mathit{4.0}\) & \(\mathbf{30.0}_{\scoreup{+7.1}}\) & \(\mathbf{86.7}\) & \(\mathit{76.7}\) & \(\mathit{76.7}\) & \(\mathit{53.3}\) & \(14.0\) & \(\mathit{55.9}_{\scoreup{+5.3}}\) \\
\hline
\multirow{3}{*}{3} & ProgCo & \(40.0\) & \(30.0\) & \(43.3\) & \(16.7\) & \(6.0\) & \(24.7_{\scoreup{+1.8}}\) & \(73.3\) & \(83.3\) & \(76.7\) & \(53.3\) & \(24.0\) & \(57.6_{\scoreup{+7.0}}\) \\
 & ProgCo-Py & \(46.7\) & \(26.7\) & \(43.3\) & \(13.3\) & \(6.0\) & \(24.7_{\scoreup{+1.8}}\) & \(76.7\) & \(73.3\) & \(80.0\) & \(60.0\) & \(28.0\) & \(59.4_{\scoreup{+8.8}}\) \\
 & \textsc{AMTFV} & \(\mathbf{53.3}\) & \(\mathbf{40.0}\) & \(\mathbf{50.0}\) & \(\mathbf{20.0}\) & \(\mathit{2.0}\) & \(\mathbf{29.4}_{\scoreup{+6.5}}\) & \(\mathbf{86.7}\) & \(\mathbf{83.3}\) & \(\mathbf{83.3}\) & \(\mathbf{60.0}\) & \(20.0\) & \(\mathbf{61.2}_{\scoreup{+10.6}}\) \\
\hline
\end{tabular}
\caption{Supplementary results for verification-enhanced methods on GPT models. The table reports detailed results for ProgCo, ProgCo-Py, and \textsc{AMTFV} with \(\texttt{max\_turn}=1\) and \(\texttt{max\_turn}=3\). Avg denotes sample-weighted average accuracy across the five datasets, and subscripts in the Avg columns show changes relative to Initial Score.}
\label{tab:appendix_gpt_progco_series_combined}
\end{table}

\begin{table}[H]
\centering
\small
\setlength{\tabcolsep}{3pt}
\renewcommand{\arraystretch}{1.08}
\begin{tabular}{llcccccccccccc}
\hline
Turn & Method & \multicolumn{6}{c}{Gemini-2.5-Flash-Lite} & \multicolumn{6}{c}{Gemini-2.5-Flash-Lite-Think} \\
\cline{3-14}
& & A24 & A25 & B25 & HMM & AMO & Avg & A24 & A25 & B25 & HMM & AMO & Avg \\
\hline
\multirow{3}{*}{1} & ProgCo & \(63.3\) & \(50.0\) & \(63.3\) & \(30.0\) & \(6.0\) & \(38.2_{\scoreup{+1.7}}\) & \(56.7\) & \(46.7\) & \(60.0\) & \(43.3\) & \(8.0\) & \(38.8_{\scoreup{+1.2}}\) \\
 & ProgCo-Py & \(63.3\) & \(46.7\) & \(63.3\) & \(36.7\) & \(6.0\) & \(38.8_{\scoreup{+2.3}}\) & \(60.0\) & \(40.0\) & \(66.7\) & \(40.0\) & \(8.0\) & \(38.8_{\scoreup{+1.2}}\) \\
 & \textsc{AMTFV} & \(\mathbf{66.7}\) & \(\mathbf{50.0}\) & \(\mathit{56.7}\) & \(\mathit{33.3}\) & \(\mathbf{8.0}\) & \(\mathbf{38.8}_{\scoreup{+2.3}}\) & \(\mathbf{60.0}\) & \(\mathbf{50.0}\) & \(\mathbf{66.7}\) & \(33.3\) & \(\mathbf{8.0}\) & \(\mathbf{39.4}_{\scoreup{+1.8}}\) \\
\hline
\multirow{3}{*}{3} & ProgCo & \(63.3\) & \(50.0\) & \(63.3\) & \(33.3\) & \(2.0\) & \(37.6_{\scoreup{+1.1}}\) & \(60.0\) & \(46.7\) & \(63.3\) & \(43.3\) & \(8.0\) & \(40.0_{\scoreup{+2.4}}\) \\
 & ProgCo-Py & \(53.3\) & \(46.7\) & \(56.7\) & \(33.3\) & \(8.0\) & \(35.9_{\scoredown{-0.6}}\) & \(66.7\) & \(46.7\) & \(70.0\) & \(40.0\) & \(8.0\) & \(41.8_{\scoreup{+4.2}}\) \\
 & \textsc{AMTFV} & \(\mathbf{66.7}\) & \(\mathbf{53.3}\) & \(\mathit{60.0}\) & \(\mathbf{33.3}\) & \(\mathbf{8.0}\) & \(\mathbf{40.0}_{\scoreup{+3.5}}\) & \(\mathit{63.3}\) & \(\mathbf{53.3}\) & \(\mathbf{73.3}\) & \(33.3\) & \(\mathbf{10.0}\) & \(\mathbf{42.4}_{\scoreup{+4.8}}\) \\
\hline
\end{tabular}
\caption{Supplementary results for verification-enhanced methods on Gemini models. The table reports detailed results for ProgCo, ProgCo-Py, and \textsc{AMTFV} with \(\texttt{max\_turn}=1\) and \(\texttt{max\_turn}=3\). Avg denotes sample-weighted average accuracy across the five datasets, and subscripts in the Avg columns show changes relative to Initial Score.}
\label{tab:appendix_gemini_progco_series_combined}
\end{table}

Overall, these supplementary results agree with the progressive architectural comparison in the main text: \textsc{AMTFV} achieves higher average accuracy than ProgCo and ProgCo-Py under most GPT and Gemini settings, indicating that the observed gains of MTF-driven verification and correction extend beyond the primary DeepSeek evaluations. The magnitude of the gains varies across models and thinking/non-thinking settings, suggesting that base-model capability and reasoning mode can affect the practical effectiveness of multistep verification and correction.

\subsection{Statistical Significance}\label{app:statistical_significance}

We conduct a paired statistical analysis using the results from seven base-model settings on the same 170 problems. \textsc{AMTFV} improves average accuracy over ProgCo-Py, the baseline with the highest overall average accuracy, by 4.03 percentage points. Using 200,000 BCa bootstrap resamples, we obtain a 95\% confidence interval of [2.10, 5.97] percentage points for this improvement. The paired permutation test yields \(p<0.001\). These results support \textsc{AMTFV}'s overall accuracy improvement under the evaluated settings.

\subsection{Iteration Budget Analysis}\label{app:iteration_budget}

\textsc{AMTFV} progressively updates either the candidate answer or the verification workflow through a multistep verification--correction loop. We analyze the effect of the maximum iteration count \(\texttt{max\_turn}\), focusing on \(\texttt{max\_turn}=1\) and \(\texttt{max\_turn}=3\). All methods are evaluated under the same maximum iteration count, so the results reflect their correction effectiveness under a fixed iteration budget.

\begin{figure}[!ht]
\centering
\includegraphics[width=\textwidth]{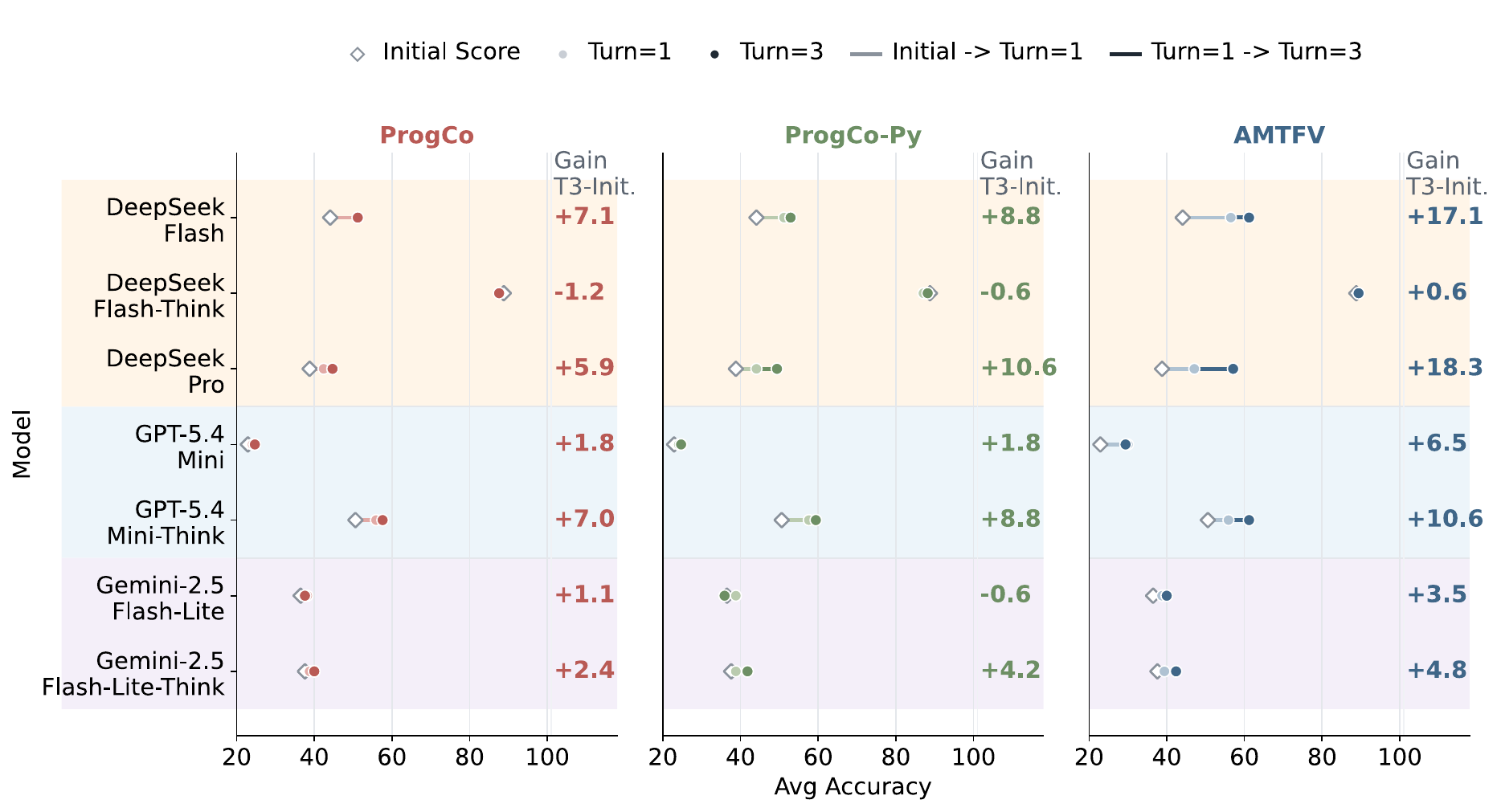}
\caption{Verification--correction performance under different iteration budgets. The three panels correspond to ProgCo, ProgCo-Py, and \textsc{AMTFV}. The horizontal axis shows sample-weighted average accuracy (\%) across the five datasets, and the vertical axis lists base-model settings. Each trajectory connects Initial Score, \(\texttt{max\_turn}=1\), and \(\texttt{max\_turn}=3\). Gain on the right denotes the absolute change from Initial Score to \(\texttt{max\_turn}=3\), in percentage points.}
\label{fig:app_iteration_budget}
\end{figure}

Figure~\ref{fig:app_iteration_budget} shows the average-accuracy trajectories of ProgCo, ProgCo-Py, and \textsc{AMTFV} under different iteration budgets. Initial Score is the accuracy before additional correction, while Turn=1 and Turn=3 correspond to \(\texttt{max\_turn}=1\) and \(\texttt{max\_turn}=3\), respectively.

From \(\texttt{max\_turn}=1\) to \(\texttt{max\_turn}=3\), \textsc{AMTFV} improves further under most base-model settings; the largest additional gain is \(+10.0\) on DeepSeek-Pro. Additional iterations are not always monotonically beneficial: on GPT-5.4-Mini, \textsc{AMTFV} decreases slightly by \(-0.6\) from \(\texttt{max\_turn}=1\) to \(\texttt{max\_turn}=3\). Multistep correction can therefore create new opportunities to fix errors but may also cause overcorrection in a small number of cases, motivating comparisons of actual correction quality under the same number of rounds.

Combined with the overall average trajectories reported in the main paper, the results indicate that \textsc{AMTFV} improves by \(+8.8\) points on average from Initial Score to \(\texttt{max\_turn}=3\), exceeding the \(+3.4\) gain of ProgCo and the \(+4.7\) gain of ProgCo-Py. \textsc{AMTFV} also continues to improve from \(\texttt{max\_turn}=1\) to \(\texttt{max\_turn}=3\), whereas the later gains of ProgCo and ProgCo-Py are smaller. Under these evaluated settings, \textsc{AMTFV} therefore appears to convert additional verification--correction rounds into effective gains more successfully than the ProgCo family.

\FloatBarrier

\subsection{Computational Cost}\label{app:computational_cost}

We additionally evaluate computational cost on 170 problems using DeepSeek-Flash in non-thinking mode. The reported costs cover verification and correction after the shared initial response, excluding initial-response generation and external final-answer evaluation. Table~\ref{tab:app_computational_cost} reports per-problem means.

\begin{table}[H]
\centering
\small
\setlength{\tabcolsep}{6pt}
\renewcommand{\arraystretch}{1.08}
\begin{tabular}{lrr}
\toprule
Metric & ProgCo-Py & \textsc{AMTFV} \\
\midrule
LLM calls & 21.96 & 21.32 \\
Total tokens (millions) & 1.078 & 0.200 \\
Python executions & 11.67 & 4.84 \\
Task completion time (min) & 20.2 & 4.0 \\
\bottomrule
\end{tabular}
\caption{Per-problem computational cost on 170 problems using DeepSeek-Flash in non-thinking mode.}
\label{tab:app_computational_cost}
\end{table}

Under the evaluated configuration, \textsc{AMTFV} has lower costs than ProgCo-Py across these metrics, with particularly substantial reductions in token consumption and Python executions. We control costs by explaining tool uses to guide selection, reminding the model when it approaches its call limit, and reusing successful results for identical requests. In this experiment, 111 cache hits avoid repeated Python execution. Task completion time is measured from each problem's first model or tool request to the final returned result, including waiting time.

\section{Verification-Complexity and Tool-Type Analysis}

\subsection{Empirical Difficulty and Verification Complexity}\label{app:difficulty_complexity}

The main text uses the instance-level number of MTF tool calls to define verification-complexity bins. We further analyze the relationship between problem-level verification complexity and empirical difficulty to assess the interpretability of this proxy signal.

Empirical difficulty is estimated from the initial-answer accuracy across multiple base-model settings:

\begin{equation}
\label{eq:empirical-difficulty}
D_{\mathrm{emp}}(q)
=
1-\frac{1}{M_{\mathrm{model}}}
\sum_{m=1}^{M_{\mathrm{model}}}
\mathbb{I}[a^0_{m,q}=a_q^*],
\end{equation}

where \(M_{\mathrm{model}}\) is the number of base-model settings, \(a^0_{m,q}\) is the initial candidate answer produced by setting \(m\) for problem \(q\), and \(a_q^*\) is the gold answer. We define average problem-level verification complexity as

\begin{equation}
\label{eq:verification-complexity}
C_{\mathrm{ver}}(q)
=
\frac{1}{M_{\mathrm{model}}}
\sum_{m=1}^{M_{\mathrm{model}}}
\log\!\left(1+n_{\mathrm{tool}}^{(m,q)}\right),
\end{equation}

where \(n_{\mathrm{tool}}^{(m,q)}\) is the number of MTF tool calls produced by \textsc{AMTFV} for model setting \(m\) and problem \(q\). The transformation \(\log(1+n_{\mathrm{tool}}^{(m,q)})\) reduces the influence of extreme call counts on the problem-level average.

The instance-level verification-complexity bins are defined as
\begin{equation}
\label{eq:complexity-bin}
b(m,q)=
\begin{cases}
\mathrm{Low}, & n_{\mathrm{tool}}^{(m,q)}\leq1,\\
\mathrm{Medium}, & 2\leq n_{\mathrm{tool}}^{(m,q)}\leq3,\\
\mathrm{High}, & n_{\mathrm{tool}}^{(m,q)}\geq4.
\end{cases}
\end{equation}

The correlation analysis in Figure~\ref{fig:app_difficulty_complexity} aggregates over 170 problems. Pearson \(r=0.67\) and Spearman \(\rho=0.70\) indicate that problems that are more difficult for the base models generally require more complex MTF verification. This analysis does not treat tool-call count as a human difficulty label; rather, it uses the count as a complexity signal from the perspective of the verification process.

\subsection{MTF Tool-Type Analysis}\label{app:mtf_tool_types}

Figure~\ref{fig:app_mtf_tool_types} shows the distribution of MTF call types in \textsc{AMTFV}. Calls are concentrated primarily in the Enumeration and Symbolic categories, indicating that exhaustive enumeration and symbolic verification are common tool requirements in hard-test mathematical tasks. This pattern is consistent with our case analyses: many incorrect answers satisfy some local conditions but omit exhaustive enumeration, global minimality, exact computation, or symbolic-equivalence checks.

\begin{center}
\begin{minipage}[t]{0.485\textwidth}
\centering
\includegraphics[width=\linewidth]{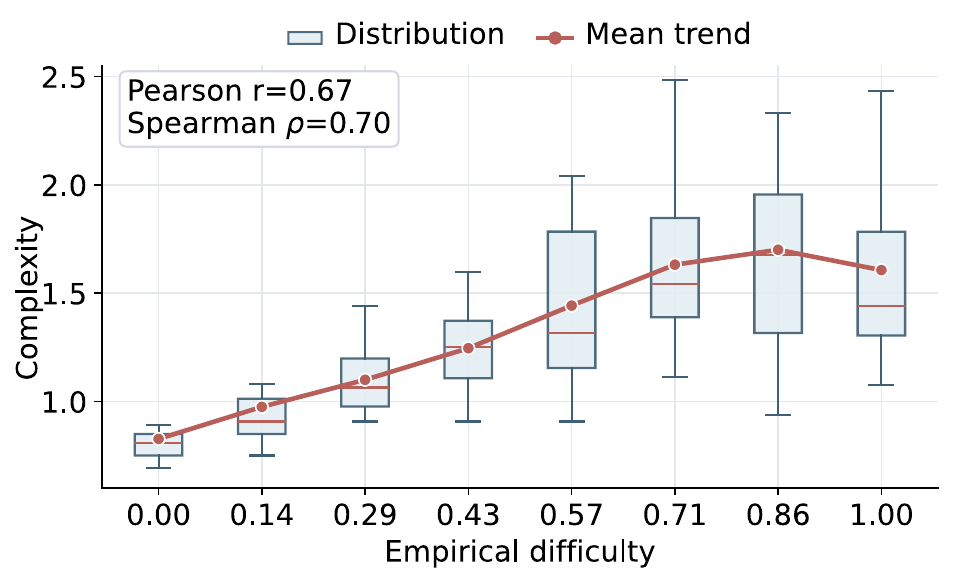}
\captionof{figure}{Relationship between empirical difficulty and verification complexity. The horizontal axis shows empirical difficulty \(D_{\mathrm{emp}}(q)\), estimated from initial-answer accuracy across seven base-model settings; the vertical axis shows average problem-level verification complexity \(C_{\mathrm{ver}}(q)\). Box plots show the distribution of problem-level verification complexity at each empirical-difficulty level, and red points and lines show the corresponding mean trend.}
\label{fig:app_difficulty_complexity}
\end{minipage}\hfill
\begin{minipage}[t]{0.485\textwidth}
\centering
\includegraphics[width=\linewidth]{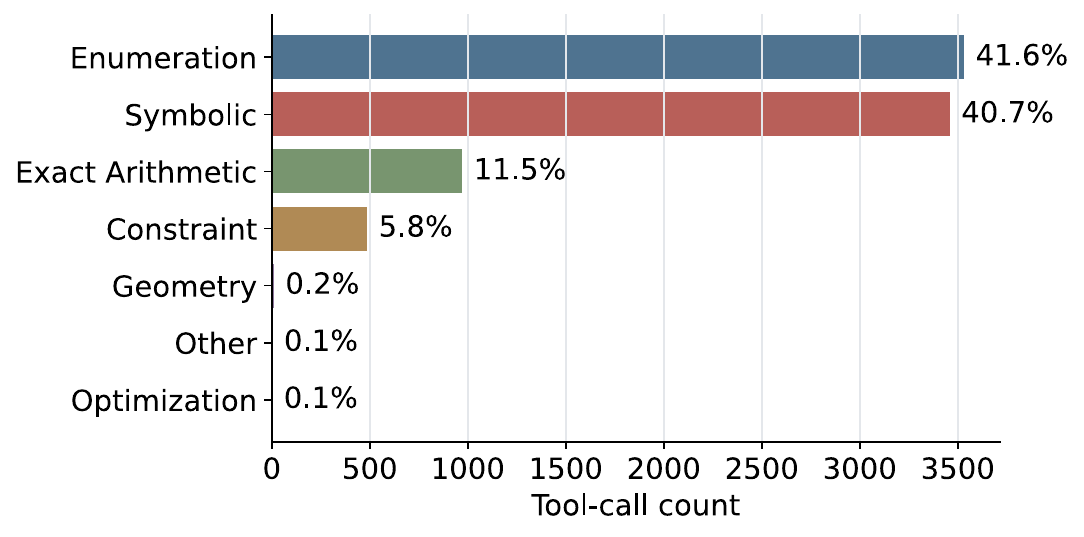}
\captionof{figure}{Distribution of MTF call types. The vertical axis lists MTF tool-call types, and the horizontal axis shows the corresponding call counts. Percentages beside the bars indicate each type's share of all MTF calls.}
\label{fig:app_mtf_tool_types}
\end{minipage}
\end{center}

These results further illustrate that \textsc{AMTFV} is designed not simply to let the model generate an arbitrary program. Instead, MTF organizes the verification target into an explicit mathematical tool-call request: the model specifies the mathematical objects, constraints, and computational objective to be verified, while the backend performs the corresponding enumeration, symbolic simplification, or exact computation.

\section{Case Studies and Full Process Comparison}

\subsection{Successful Case Study}\label{app:case_study}

To illustrate how \textsc{AMTFV} uses MTF and mathematical tool results for correction, we compare several representative successful cases. In every case, the initial response is incorrect, neither ProgCo nor ProgCo-Py corrects it, and \textsc{AMTFV} obtains the correct answer through MTF-driven verification and correction. Supplementary Table~\ref{tab:successful_cases} summarizes the key differences. The next section presents the BRUMO25 exact double-summation case in full as a direct illustration of how \textsc{AMTFV} models a mathematical verification target as an exact tool computation and uses structured tool results for candidate-answer adjudication and revision.

\begin{center}
\begin{minipage}{\textwidth}
\centering
\small
\setlength{\tabcolsep}{4pt}
\begin{tabularx}{\textwidth}{lccccX}
\toprule
Case & Initial & ProgCo & ProgCo-Py & \textsc{AMTFV} & Key MTF role \\
\midrule
AIME24 minimality & 155 & 155 & 155 & 110 & Verify global minimality \\
AIME25 counting & 764 & 764 & 764 & 16 & Exhaustively enumerate valid cases \\
AMO lattice triangles & 7 & 7 & 7 & 6 & Enumerate 15 points; maximum isosceles-free subset: 5 \\
BRUMO25 exact double sum & \(4608\) & \(4608\) & \(4608\) & \(4529\) & Compute the exact sum and return \(S,A,R\) \\
\bottomrule
\end{tabularx}
\captionof{table}{Representative successful correction cases. The table reports the initial answer, baseline outputs, and \textsc{AMTFV}'s final answer.}
\label{tab:successful_cases}
\end{minipage}
\end{center}

\textbf{AIME24 minimality verification.}
The problem asks for the smallest value satisfying the given conditions. The initial response gives 155, which satisfies some conditions but is not the required minimum; neither ProgCo nor ProgCo-Py corrects it. In contrast, \textsc{AMTFV} not only checks feasibility but also constructs a minimality target and invokes mathematical tools through MTF to search for a smaller valid candidate. The tool result supports rejecting 155, and the system ultimately revises the answer to 110.

\textbf{AIME25 combinatorial counting.}
The initial response considers only a subset of the valid cases and returns 764; ProgCo and ProgCo-Py retain the same incorrect answer. \textsc{AMTFV} instead expresses the counting target as an exhaustive MTF enumeration task, asking the backend to enumerate all combinations satisfying the original constraints. The tool result exposes cases omitted by the initial reasoning, triggers answer revision, and yields the correct answer 16.

\textbf{AMO lattice-point isosceles triangles.}
The problem asks for the smallest \(n\) such that every selection of \(n\) points from 15 triangular-lattice points contains an isosceles triangle. The initial response, ProgCo, and ProgCo-Py all return \(7\), relying primarily on the incorrect assumption that the largest isosceles-free subset has size 6. \textsc{AMTFV} represents the 15 lattice points as
\[
P=\{(i,j):0\leq j\leq4,\ 0\leq i\leq4-j\},
\]
and exhaustively checks all 6-point subsets. The tool result shows that every 6-point subset contains an isosceles triangle, while the following 5-point counterexample exists:
\[
\{(0,0),(1,0),(2,0),(3,0),(4,0)\},
\]
Therefore, the largest isosceles-free subset has size 5, and the smallest guaranteed number is \(6\). This case illustrates the role of MTF in exhaustive enumeration over a finite geometric structure and verification of a global guarantee.

\textbf{BRUMO25 exact double summation.}
The initial response, ProgCo, and ProgCo-Py all accept the candidate answer \(4608\); their common error is assuming that the sum of modular-power residues is \(4656\) for every fixed \(n\). \textsc{AMTFV} expresses the verification target as an exact double-summation task and asks the tool to return the total residue \(S\), exact value \(A\), and rounded result \(R\). The tool returns \(S=439312\), \(A=439312/97\), and \(R=4529\), thereby rejecting 4608 and correcting the answer.

These cases illustrate two characteristic advantages of \textsc{AMTFV}. First, MTF extends candidate verification to global constraints, including minimality, completeness, and feasible-region coverage. Second, it delegates enumeration, symbolic computation, and exact numerical computation---steps easily omitted in natural-language reasoning---to the mathematical toolbox backend. By writing tool results to the execution record, the system can further support subsequent adjudication, answer revision, and verification-workflow revision.

\subsection{Full Case Process Comparison: Exact Double-Sum Verification}
\label{app:case_exact_sum}

We examine a BRUMO25 exact double-summation case to illustrate how Initial, ProgCo, ProgCo-Py, and \textsc{AMTFV} differ in verification-target construction, tool execution, and use of feedback. The problem asks for
\[
A=\sum_{a=1}^{96}\sum_{n=1}^{96}\left\{\frac{a^n}{97}\right\}
\]
rounded to the nearest integer. The gold answer is \(4529\), whereas Initial, ProgCo, and ProgCo-Py all return \(4608\).

The common error in the first three outputs is the following implicit assumption: for every fixed \(n\), as \(a\) ranges from \(1\) to \(96\),
\[
\sum_{a=1}^{96}(a^n\bmod 97)=1+2+\cdots+96=4656.
\]
This assumption holds only when \(a\mapsto a^n\pmod{97}\) is a permutation of \((\mathbb Z/97\mathbb Z)^\times\), which requires \(\gcd(n,96)=1\). When \(\gcd(n,96)>1\), the image is only a proper subgroup, so the inner sum need not equal \(4656\). The central issue is therefore not a simple arithmetic error but an incorrect group-structure assumption hidden in candidate verification.

At the programmatic-verification level, the methods differ primarily in how they organize the computational target and returned result. ProgCo generates double-loop verification code based on modular exponentiation, but its final feedback remains primarily a Boolean verdict. ProgCo-Py attempts to compute the fractional part of \(a^n/97\) directly, coupling tool execution with huge powers and floating-point precision. \textsc{AMTFV}, in contrast, explicitly represents the target as three mathematical objects, \(S,A,R\), and asks the tool to return exact results together with candidate-matching status. We next compare how the methods process this candidate answer.

\begin{breakablecasebox}
\small
\textbf{Problem, ground truth, and Initial response.}

\textbf{Problem.}
Compute
\[
A=\sum_{a=1}^{96}\sum_{n=1}^{96}\left\{\frac{a^n}{97}\right\},
\]
where \(\{x\}\) denotes the fractional part of \(x\), and return \(A\) rounded to the nearest integer.

\textbf{Ground truth.}
The gold answer is
\[
4529.
\]

\textbf{Initial extracted answer.}
The final answer extracted from Initial is
\[
4608.
\]

\textbf{Generated solution.}
Initial first rewrites each term as
\[
\left\{\frac{a^n}{97}\right\}
=
\frac{a^n\bmod97}{97},
\]
and hence
\[
A=
\frac{1}{97}
\sum_{a=1}^{96}\sum_{n=1}^{96}(a^n\bmod97).
\]
Initial then assumes that, for each fixed \(n\), the values of \(a^n\bmod97\) as \(a\) ranges from \(1\) to \(96\) can be treated as either a permutation or equal repetitions of all nonzero residue classes \(1,2,\ldots,96\). It therefore uses
\[
\sum_{a=1}^{96}(a^n\bmod97)=1+2+\cdots+96=4656
\]
and obtains
\[
A=\frac{96\cdot4656}{97}=96\cdot48=4608.
\]
The final answer is
\[
\boxed{4608}.
\]
\end{breakablecasebox}

Initial's central error is failing to distinguish \(\gcd(n,96)=1\) from \(\gcd(n,96)>1\). In the latter case, \(a\mapsto a^n\pmod{97}\) is not a permutation of the nonzero residue classes, so the inner sum need not equal \(4656\). The incorrect candidate \(4608\) therefore arises from inadequate modeling of the mathematical object rather than an error in the final numerical simplification.

\begin{breakablecasebox}
\small
\textbf{ProgCo.}

\textbf{Final extracted answer.}
\[
4608.
\]

\textbf{Generated solution.}
ProgCo's final response retains the same central mathematical assumption as Initial. It likewise writes the fractional part as
\[
\left\{\frac{a^n}{97}\right\}
=
\frac{a^n\bmod97}{97},
\]
and assumes that, for each fixed \(n\), the inner sum equals the sum of all nonzero residue classes:
\[
\sum_{a=1}^{96}(a^n\bmod97)=4656.
\]
ProgCo therefore obtains
\[
A=\frac{96\cdot4656}{97}=4608,
\]
and returns the final answer
\[
\boxed{4608}.
\]

\textbf{Generated verification code.}
ProgCo further generates a verification function for the candidate answer \(4608\). The function attempts to compute the original double sum directly:
\begin{flushleft}
\ttfamily\footnotesize
verify\_fractional\_part\_sum(answer):\\
\hspace*{1em}A = 0\\
\hspace*{1em}for a in range(1, 97):\\
\hspace*{2em}for n in range(1, 97):\\
\hspace*{3em}remainder = pow(a, n, 97)\\
\hspace*{3em}A += remainder / 97\\
\hspace*{1em}rounded\_A = round(A)\\
\hspace*{1em}return rounded\_A == answer
\end{flushleft}
The verification code computes each remainder using modular exponentiation:
\[
\texttt{remainder = pow(a,n,97)}
\]
and accumulates
\[
\texttt{A += remainder / 97}.
\]
Its intended computation is
\[
A=\frac{1}{97}\sum_{a=1}^{96}\sum_{n=1}^{96}(a^n\bmod97),
\]
and then to compare whether
\[
\operatorname{round}(A)
\]
equals the candidate answer \(4608\).

\textbf{Returned verification feedback.}
The retained verification feedback states that the \(96\times96=9216\) double loop can be computed directly and claims that the code is independent of the original reasoning and will produce a result consistent with the candidate. This round therefore returns
\[
\texttt{True},
\]
with feedback
\[
\texttt{verify passed}.
\]
Notably, the feedback does not return the computed total residue \(S\), exact value \(A\), or rounded result \(R\); it accepts the candidate through a Boolean verdict.
\end{breakablecasebox}

ProgCo's process reflects program-driven verification: beyond producing a candidate answer, it attempts to express verification as a finite-summation program. This design is more concrete than natural-language reflection alone and seeks to turn answer verification into an executable process.

In this case, however, ProgCo's verification feedback does not explicitly return the key mathematical objects
\[
\begin{aligned}
S&=\sum_{a=1}^{96}\sum_{n=1}^{96}(a^n\bmod97),\\
A&=\frac{S}{97},\\
R&=\operatorname{round}(A).
\end{aligned}
\]
Instead, the feedback only reports \(\texttt{verify passed}\) and accepts \(4608\). Although ProgCo generates an apparently direct verification function, its tool feedback does not expose the erroneous assumption in Initial as a traceable, exact mathematical result.

\begin{breakablecasebox}
\small
\textbf{ProgCo-Py.}

\textbf{Final extracted answer.}
\[
4608.
\]

\textbf{Generated solution.}
ProgCo-Py's final response likewise retains Initial's core reasoning by transforming
\[
\left\{\frac{a^n}{97}\right\}
\]
into
\[
\frac{a^n\bmod97}{97},
\]
and assuming that the inner sum for every \(n\) is
\[
1+2+\cdots+96=4656.
\]
It therefore still obtains
\[
A=\frac{96\cdot4656}{97}=4608,
\]
and returns
\[
\boxed{4608}.
\]

\textbf{Generated Python verification code.}
ProgCo-Py generates Python-style direct-computation code to verify \(4608\). Unlike ProgCo's modular-power implementation, this code first attempts to compute \(a^n/97\) and then extract its fractional part:
\begin{flushleft}
\ttfamily\footnotesize
verify\_fractional\_part\_sum(answer):\\
\hspace*{1em}A = 0\\
\hspace*{1em}for a in range(1, 97):\\
\hspace*{2em}for n in range(1, 97):\\
\hspace*{3em}value = (a ** n) / 97\\
\hspace*{3em}fractional\_part = value - int(value)\\
\hspace*{3em}A += fractional\_part\\
\hspace*{1em}rounded\_A = round(A)\\
\hspace*{1em}return rounded\_A == answer
\end{flushleft}
That is, ProgCo-Py's verification code directly computes
\[
\texttt{value = (a ** n) / 97}
\]
and obtains the fractional part via
\[
\texttt{fractional\_part = value - int(value)}
\]
.

\textbf{Python execution and returned result.}
The execution feedback notes that directly computing \(a^n/97\) involves extremely large integers \(a^n\), which may cause floating-point overflow or severe precision loss; the direct floating-point implementation is therefore unreliable. The verification process then falls back to checking the mathematical reasoning in the original solution and again accepts the claim
\[
\sum_{a=1}^{96}(a^n\bmod97)=4656
\]
for all \(n\). This round consequently retains
\[
A=4608,
\]
The verification result is
\[
\texttt{True},
\]
with feedback
\[
\texttt{verify passed}.
\]
\end{breakablecasebox}

ProgCo-Py's process illustrates that adding Python-tool feedback does not automatically ensure reliable verification. The problem is not the absence of a tool call, but the continued coupling between tool implementation and the mathematical verification target. The code computes the fractional part of \(a^n/97\) directly and is therefore vulnerable to huge powers and floating-point precision; once direct execution is deemed infeasible, verification returns to the original incorrect assumption. In other words, the tool call is not reliably converted into an exact mathematical-object computation and thus fails to correct the candidate \(4608\).

\begin{breakablecasebox}
\small
\textbf{\textsc{AMTFV}.}

\textbf{Final extracted answer.}
\[
4529.
\]

\textbf{Initial candidate.}
\textsc{AMTFV} starts from the same initial candidate answer:
\[
4608.
\]

\textbf{MTF verification target.}
Rather than asking the model to judge the plausibility of \(4608\) directly or relying on the group-theoretic assumption in the original solution, the verification agent expresses the problem requirement as an exact mathematical computation target:
\[
\begin{aligned}
S&=\sum_{a=1}^{96}\sum_{n=1}^{96}(a^n\bmod97),\\
A&=\frac{S}{97},\\
R&=\operatorname{round}(A).
\end{aligned}
\]
The candidate passes verification if and only if
\[
R=4608.
\]

The corresponding MTF request can be summarized as follows:
\begin{flushleft}
\ttfamily\footnotesize
<tool\_flow>\\
\hspace*{1em}Let p = 97.\\
\hspace*{1em}Compute S = sum\_\{a=1\}\^{}96 sum\_\{n=1\}\^{}96 (a\^{}n mod p).\\
\hspace*{1em}Compute A = S / p.\\
\hspace*{1em}Compute R = round(A).\\
\hspace*{1em}Return S, A, R, and whether R == 4608.\\
</tool\_flow>
\end{flushleft}

\textbf{Compiled tool call and structured return.}
The mathematical toolbox agent translates this MTF request into exact integer and rational arithmetic rather than direct computation with huge floating-point values. The mathematical toolbox backend accumulates integer residues using modular exponentiation:
\[
S=\sum_{a=1}^{96}\sum_{n=1}^{96}\operatorname{pow}(a,n,97),
\]
It then computes, in exact rational form,
\[
A=\frac{S}{97}.
\]
The tool returns the following structured result:
\[
\begin{aligned}
S&=439312,\\
A&=\operatorname{Fraction}(439312,97),\\
R&=4529,\\
(R=4608)&=\texttt{False}.
\end{aligned}
\]

\textbf{Tool execution result.}
The tool result gives
\[
A=\frac{439312}{97}
=
4528+\frac{96}{97},
\]
and hence
\[
R=\operatorname{round}(A)=4529.
\]
The result also reports
\[
R=4529\neq4608.
\]
Therefore, the candidate answer \(4608\) fails verification:
\[
\texttt{False}.
\]

\textbf{Feedback and answer revision.}
Based on the tool result, the feedback identifies Initial's key error: it assumes that for every fixed \(n\), the map
\[
a\mapsto a^n\pmod{97}
\]
covers all nonzero residue classes. In fact, this map is a permutation only when \(\gcd(n,96)=1\); when \(\gcd(n,96)>1\), its image is a proper subgroup and the inner sum need not equal \(4656\).

Using this feedback, the answer-revision stage recomputes the double sum and obtains
\[
A=\frac{439312}{97}
=
4528+\frac{96}{97}.
\]
The rounded result is therefore
\[
4529.
\]

\textbf{Final verification.}
Final verification uses
\[
4529
\]
as the candidate answer and invokes the same exact-summation verification target again:
\[
\begin{aligned}
S&=\sum_{a=1}^{96}\sum_{n=1}^{96}(a^n\bmod97),\\
A&=\frac{S}{97},\\
R&=\operatorname{round}(A).
\end{aligned}
\]
The tool again returns
\[
S=439312,\qquad
R=4529.
\]
Final verification therefore succeeds:
\[
\texttt{True},
\qquad
\texttt{verify passed}.
\]
\textsc{AMTFV} ultimately returns
\[
\boxed{4529}.
\]
\end{breakablecasebox}

A key advantage of \textsc{AMTFV} in this case is that it neither asks the model to assess the original reasoning in natural language nor asks it to generate ad hoc code that determines how to compute the answer. Instead, it first models the verification target explicitly as the exact mathematical objects
\[
\begin{aligned}
S&=\sum_{a=1}^{96}\sum_{n=1}^{96}(a^n\bmod97),\\
A&=\frac{S}{97},\\
R&=\operatorname{round}(A).
\end{aligned}
\]
The mathematical toolbox backend then performs the finite summation, division by \(97\), rounding, and candidate comparison. In this case, the process decouples mathematical modeling from concrete execution: the verification agent specifies \emph{what to compute}, the backend performs \emph{how to compute it} using exact arithmetic, and the returned results support candidate-answer adjudication and revision.

\paragraph{Discussion.}
The key distinction in this case is not whether code is used, but whether code or tool calls serve an explicit, traceable mathematical verification target. ProgCo and ProgCo-Py both attempt to program candidate verification, yet their feedback does not reliably produce exact mathematical objects that support adjudication and correction. In contrast, \textsc{AMTFV} organizes the target as an MTF request to the mathematical toolbox, causing the tool to return \(S\), \(A\), \(R\), and whether the candidate matches. The system can therefore directly identify the discrepancy between \(4608\) and the true rounded result \(4529\), and revise the candidate to the correct answer.
 
\end{document}